\documentclass[sigconf,natbib=false]{acmart}

\usepackage{rotating}
\usepackage{xcolor}
\usepackage{graphicx}

\AtBeginDocument{%
  \providecommand\BibTeX{{%
    \normalfont B\kern-0.5em{\scshape i\kern-0.25em b}\kern-0.8em\TeX}}}

\setcopyright{acmcopyright}
\copyrightyear{2018}
\acmYear{2018}
\acmDOI{XXXXXXX.XXXXXXX}

\acmPrice{XX.XX}
\acmISBN{978-1-4503-XXXX-X/18/06}

\newtheorem{property}{Property}

\begin{document}

%\title{Formal Concept Analysis: a structural framework naturally embedding variability (8+2p)}
%\title{Formal Concept Analysis and its conceptual structures: a natural variability analysis framework}
\title{Formal Concept Analysis: a Structural Framework for Variability Extraction and Analysis}

\author{Jessie Galasso}
\email{jessie.galasso-carbonnel@mcgill.ca}
\orcid{0000-0002-9868-1814}
\affiliation{%
  \institution{McGill University}
  \country{Canada}
}


\begin{abstract}
Formal Concept Analysis (FCA) is a mathematical framework for knowledge representation and discovery.
It performs a hierarchical clustering over a set of objects described by attributes, resulting in conceptual structures in which objects are organized depending on the attributes they share.
These conceptual structures naturally highlight commonalities and variabilities among similar objects by categorizing them into groups which are then arranged by similarity, making it particularly appropriate for variability extraction and analysis.
Despite the potential of FCA, determining which of its properties can be leveraged for variability-related tasks (and how) is not always straightforward, partly due to the mathematical orientation of its foundational literature.
This paper attempts to bridge part of this gap by gathering a selection of properties of the framework which are essential to variability analysis, and how they can be used to interpret diverse variability information within the resulting conceptual structures.
\end{abstract}


\maketitle

\section{Introduction}
\label{sec:introduction}

% Definitions and scope of variability

The concept of \textit{variability} is the cornerstone of software product line engineering~\cite{pohl2005software}, where variability modelling consist in representing what is common and what varies within a software system family~\cite{capilla2013systems}.
Variability management gathers methods and tools promoting systematic reuse of commonalities and control over variabilities, as well as mass customization.
This approach, centred around planned reuse, is flexible enough to help improve industrial practice regarding management of software variants without going to the full extent of implementing a product line, a challenge identified in 2014 by Metzger and Pohl as “variability management in non-product line settings”~\cite{metzger2014software}.
We can thus see variability management as a spectrum with opportunistic reuse in one end and a full product line in the other end.
In fact, the definition and principles behind variability management are generic enough to be applied in a wide range of domains, as demonstrated by works on user interface design~\cite{martinez2017variability}, data visualization~\cite{horcas2022variability}, 3D modelling~\cite{jacobs2023varimod} or computational notebooks~\cite{brault2023taming}. 

Bringing variability management in settings where variability was not planned in advance necessitates as a first step to elicit and formalize variability information from existing variants.
This is a non-trivial task which greatly benefits from automated variability extraction methods and tools~\cite{lopez2022handbook}.
In this work, we focus on variability extraction at large, as a key step for bringing  variability management in existing variants of software systems, or of different kinds of  artifacts which do not qualify as a software family \textit{per se} but are similar enough to benefit from such methodologies.

% Introduction to FCA and link to variaiblity

Formal Concept analysis (FCA)~\cite{ganter1999formal} is a mathematical framework for data analysis and knowledge extraction. 
As input, it takes a set of objects described by binary attributes, and create a hierarchy of clusters organizing the objects depending on the attributes they share.
Clusters represent groups of similar objects, and the hierarchy shows clusters' step-wise division: the largest clusters are situated at the top and each subsequent level presents the progressive subdivision of upper clusters into smaller groups of objects with higher similarity.
This hierarchy both captures variants' similarity at several levels of increasing details and highlights where the variations occur.
\textbf{This framework stands out as great candidate for identifying variability information because it effectively highlights commonalities and variabilities within a set of comparable objects by classifying them into groups which are in turn organized by similarity.}
The conceptual structures produced by mean of FCA describe natural representations which are fundamental to hierarchies and object/attribute structures~\cite{priss2006formal}, which explains why these structures were re-discovered several times in different areas of application.
FCA is often referred to as yet another method for variability extraction, when in fact it can be seen as a formalization of these properties and associated representations.
The theory behind FCA was never intended to capture certain types of knowledge (it is, in this sense, not a \textit{functional approach}) but rather to formalize properties inherent to hierarchies and object/attribute structures.
This is why we argue that it a \textit{structural approach}, which appears to naturally unveil variability information by structuring objects depending on their shared attributes.

FCA is at the core of several works on variability analysis and extraction~\cite{shatnawi2017recovering,DBLP:conf/aosd/NiuE09,DBLP:conf/csmr/LoeschP07,DBLP:conf/wcre/YangPZ99a,DBLP:conf/cla/Al-MsiedeenHSUV14,DBLP:conf/splc/Georges22,DBLP:conf/vamos/GeorgesRHKNT23,DBLP:journals/jss/CarbonnelHN19,DBLP:conf/wcre/XueXJ12,DBLP:conf/gpce/HladLHS21} in which the framework is predominantly utilized as a knowledge extraction tool to address specific problems.
Each one of these works usually relies on a subset of FCA properties and focuses on its practical application, giving less priority to the theoretical aspects of the framework and how they map to variability in general. Although this collection of works showcases diversified usages of FCA in the context of variability analysis and extraction, a comprehensive overview of the links between the different framework’s properties and how they map to variability information is not readily apparent. The mathematical nature of its foundational literature further complicates determining FCA's relevance for a given problem and how it can be applied.
% OBJECTIVES AND CONTRIBUTIONS
The objective of this paper is to consolidate information on the FCA process and its mapping to variability extraction: it aims at offering a comprehensive guide about the identification of the diverse variability information contained in the resulting conceptual structures. 
To this end, it first presents the core definitions of FCA by detailing step-by-step the process of building conceptual structures with an illustrative example in Sect.~\ref{sec:definitions}. Then, it dissects the resulting conceptual structure along different perspectives to show how to interpret the variability information naturally present in it in Sect.~\ref{sec:properties}.
Strategies for managing scalability issues, extensions of the framework and existing tools are covered in Sect.~\ref{sec:extensions}.
Related work (Sect.~\ref{sec:rw}) gathers studies which employ FCA for variability extraction tasks: it shows that the same framework supports a wide range of tasks and applies to diverse artifacts without the necessity to be adapted.

The main contribution of this paper is to bring together established properties commonly used for variability extraction, along with some which were, to the best of our knowledge, never linked to this task. 
Although not introducing novel methodologies, this work serves as a starting point for practitioners and researchers interested in using FCA for variability extraction.
\section{Definitions}
\label{sec:definitions}

In this section, we define and illustrate the core steps of the process of FCA.
FCA is a mathematical framework building a hierarchy of clusters from a set of objects described by attributes.
As input, it takes a\textit{ formal context} (Sect.~\ref{sec:formalcontext}) depicting objects and their attributes.
FCA theory  relies on the definition of two \textit{derivation operators} (Sect.~\ref{sec:operators}) which are applied on the formal context to identify clusters of objects called \textit{formal concepts} (Sect.~\ref{sec:formalconcepts}).
The derivation operators ensure that the formal concepts verify key properties that define a \textit{ partial order} between them (Sect.~\ref{sec:partialorder}), thus forming a hierarchy of clusters   called a \textit{concept lattice} (Sect.~\ref{sec:conceptlattice}).

\subsection{Formal Context}
\label{sec:formalcontext}

A formal context is a representation of a finite set of objects (or \textit{elements}) and the attributes (or \textit{characteristics}) they own. 
The attributes are binary, thus for each attribute, an object can own it or not.
Formally, a formal concept is a triple $K = (O, A, R)$ with $O$ the finite set of objects, $A$ the finite set of attributes, and $R \subseteq O \times A$ is a relation depicting which objects own which attributes.

A formal context can be represented by a binary table in which rows are objects and columns are attributes. 
Each pair $(o, a) \in R$ is represented by a cross in the cell corresponding to the object $o$ and the attribute $a$.
Table~\ref{tab:formalcontext} presents a formal context $K_{DM}$ inspired by a comparison matrix from Wikipedia\footnote{\url{https://en.wikipedia.org/wiki/Comparison_of_data_modeling_tools}, accessed in August 2023}.
It shows 5 data modelling tools (objects) and 7 attributes representing their supported operating systems (\texttt{OS}) among \texttt{Windows}, \texttt{Mac OS} and \texttt{Linux}, as well as their supported data models (\texttt{DM}) among \texttt{Conceptual}, \texttt{Physical}, \texttt{Logical} and \texttt{ETL}.  
For instance, in this table, we can read that:

$\bullet$ the object \textit{MySQL-Workbench} possesses the four attributes \texttt{OS:Windows}, \texttt{OS:Mac}, \texttt{OS:Linux} and \texttt{DM:Physical};

$\bullet$ the attribute \texttt{DM:Logical} is shared by the three objects \textit{Erwin-DM}, \textit{ER-Studio} and \textit{Magic-Draw}.

The set of attributes describing an object is kindred to a \textit{valid configuration} in variability and product line engineering.
Table~\ref{tab:formalcontext} thus presents 5 valid configurations  corresponding to the 5 objects.

\begin{table}[ht]
\caption{Formal context presenting 5 objects (data modelling tools) and 7 attributes (operating systems and data models)}
\label{tab:formalcontext}
\centering
\small
\begin{tabular}{l|ccccccc}
DM tools ($K_{DM}$) & 
\begin{sideways}\texttt{OS:Windows}\end{sideways} & 
\begin{sideways}\texttt{OS:Mac}\end{sideways} & 
\begin{sideways}\texttt{OS:Linux}\end{sideways} & 
\begin{sideways}\texttt{DM:Conceptual}\end{sideways} & 
\begin{sideways}\texttt{DM:Physical}\end{sideways} & 
\begin{sideways}\texttt{DM:Logical}\end{sideways} & 
\begin{sideways}\texttt{DM:ETL}\end{sideways} \\
\hline
\textit{Astah} & $\times$ & $\times$ & $\times$ & $\times$ &  &  &  \\
\textit{Erwin-DM} & $\times$ &  &  & $\times$ & $\times$ & $\times$ &  \\
\textit{ER-Studio} & $\times$ &  &  & $\times$ & $\times$ & $\times$ & $\times$ \\
\textit{Magic-Draw} & $\times$ & $\times$ & $\times$ & $\times$ & $\times$ & $\times$ &  \\
\textit{MySQL-Workbench} & $\times$ & $\times$ & $\times$ &  & $\times$ &  & 
\end{tabular}
\end{table}

\subsection{Derivation Operators}
\label{sec:operators}

FCA relies on two derivation operators named $\alpha$ and $\beta$. They define the cluster extraction process from the formal context.
$\alpha: 2^{O} \mapsto 2^{A}$ associates a subset of the objects of $O$ with all attributes shared by these objects.
Conversely, $\beta: 2^{A} \mapsto 2^{O}$ associates a subset of the attributes of $A$ with all objects sharing these attributes.
For instance, $\alpha(\{$\textit{Erwin-DM, ER-Studio}$\})$ returns $\{\texttt{OS:Windows}, \texttt{DM:Conceptual},$ $ \texttt{DM:Physical}, \texttt{DM:Logical}\}$, 
and $\beta(\{\texttt{OS:Linux}, \texttt{DM:Conceptual}\})$ returns $\{$\textit{Astah, Magic-Draw}$\}$. 
The composition of $\alpha$ and $\beta$ forms a \textit{closure operator}.
A closure operator is a mapping over a set of elements, which is extensive, increasing and idempotent.

$\beta \circ \alpha$ is a closure operator over the set of objects $O$: it takes a subset of $O$ as input, and outputs another subset of $O$.
For instance, we saw earlier that $\alpha(\{$\textit{Erwin-DM, ER-Studio}$\})$ returns $\{$\texttt{OS:Windows}, \texttt{DM:Conceptual}, \texttt{DM:Physical}, \texttt{DM:Logical}$\}$.
Now, if we apply $\beta$ on these attributes, we have $\beta(\{$\texttt{OS:Windows}, \texttt{DM:Conceptual}, \texttt{DM:Physical},  \texttt{DM:Logical} $\}) = \{$\textit{Erwin-DM, ER-Studio, Magic-Draw}$\}$.
The initial subset of objects is included in the one we obtain ($\beta \circ \alpha$ is extensive).
The obtained subset is said to be \textit{closed}, because if we reapply $\beta \circ \alpha$ on it, we re-obtain the same ($\beta \circ \alpha$ is idempotent).
We also say that this subset is \textit{maximal}, because it is the largest subset of objects sharing the same attributes.
Finally, the increasing property of the operator ensures that if $O_1 \subseteq O_2 \subseteq O$, then $\beta \circ \alpha(O_1) \subseteq \beta \circ \alpha(O_2)$.
Conversely, $\alpha \circ \beta$ is a closure operator over the set of attributes $A$. It associates each subset of $A$ to another subset of $A$ which is maximal.

\subsection{Formal Concepts}
\label{sec:formalconcepts}

The process of FCA consists of applying the two aforementioned operators on a formal context to extract a finite set of \textit{formal concepts}.
A formal concept is a bi-cluster representing a maximal set of objects sharing a maximal set of attributes.
Formally, a formal concept is a pair $C = (E, I)$, with $E \subseteq O$ and $I \subseteq A$, verifying $\alpha(E) = I$ and $\beta(I) = E$. 
$E$ is called the extent of the concept, and $I$ its intent.
According to the previous example, the pair  $(\{$\textit{Erwin-DM, ER-Studio, Magic-Draw}$\}, \{$\texttt{OS:Windows}, \texttt{DM:Conceptual}, \texttt{DM:Physical}, \texttt{DM:Logical}$ \}$) is a formal concept of $K_{DM}$.
Graphically, a formal concept corresponds to a maximal rectangle of crosses in the table, which can be obtained by permuting rows and columns.

The intent and the extent of a formal concept are \textit{closed}.
In other words, there are no other objects than the ones in $E$ which share all attributes in $I$, there are no other attributes than $I$ which are shared by all objects in $E$.
A formal concept is thus a maximal group of similar objects with all the attributes they have in common.
The application of the derivation operators guarantees that:
\begin{property}
All extracted clusters are maximal, and  all maximal clusters are extracted.     
\end{property}

The sets $O$ and $A$ share a \textit{Galois connection}, which means that if one selects more elements of one set, they correspond to fewer elements of the other set, and conversely.
A concept with numerous objects in its extent will have a small set of shared attributes.
A concept with fewer objects will conversely have more attributes in its intent.
A concept can be seen as a class in the object-oriented paradigm: the concept's intent acts as the attributes of the class, and the concept's extent as its instances. A sub-class has more attributes than its super-class: it is more specific and matches fewer instances.

\subsection{Partial Order Between Formal Concepts}
\label{sec:partialorder}

Identified formal concepts are then partially ordered by inclusion on their extent.
Given two formal concepts $C_1 = (E_1, I_1)$ and $C_2 = (E_2, I_2)$, $C_1 \leq C_2 \Leftrightarrow E_1 \subseteq E_2$.
This order is equivalent to the reverse-inclusion on their intent, i.e., $C_1 \leq C_2 \Leftrightarrow I_2 \subseteq I_1$, thanks to the Galois connection.
Let us consider the previous concept $C = (E, I)$ with $E = \{$\textit{Erwin-DM, ER-Studio, Magic-Draw}$\}$ and $I = \{$\texttt{OS:Windows}, \texttt{DM:Conceptual}, \texttt{DM:Physical},$ $  \texttt{DM:Logical}$\}$.
If we want the group of objects having these attributes plus \texttt{DM:ETL} (i.e., $I \cup \{\texttt{DM:ETL}\}$), we will naturally obtain a subset of the objects of $E$, because the latters are all the objects having the attributes of $I$. 
We can build the corresponding formal concept $C' = (E', I')$ with $E' = \alpha(I \cup \{\texttt{DM:ETL}\}) = \{ER$-$Studio\}$ and $I' = \beta(\{ER$-$Studio\}) = \{$\texttt{OS:Windows}, \texttt{DM:Conceptual}, \texttt{DM:Physical}, \texttt{DM:Logical}, \texttt{DM:ETL}$\}$.
Because $I \subseteq I'$ (and thus $E' \subseteq E$), then $C' \leq C$. 
We say that $C'$ is a \textbf{sub-concept} of $C$, and that $C$ is a \textbf{super-concept} of $C'$.
This partial order is sometimes referred to as a \textbf{specialization order} analogous to the specialization/generalization relationship between super- and sub-classes in the object-oriented paradigm.

\subsection{Concept Lattice}
\label{sec:conceptlattice}

\begin{figure}[ht]
    \centering
    \includegraphics[width=.9\linewidth]{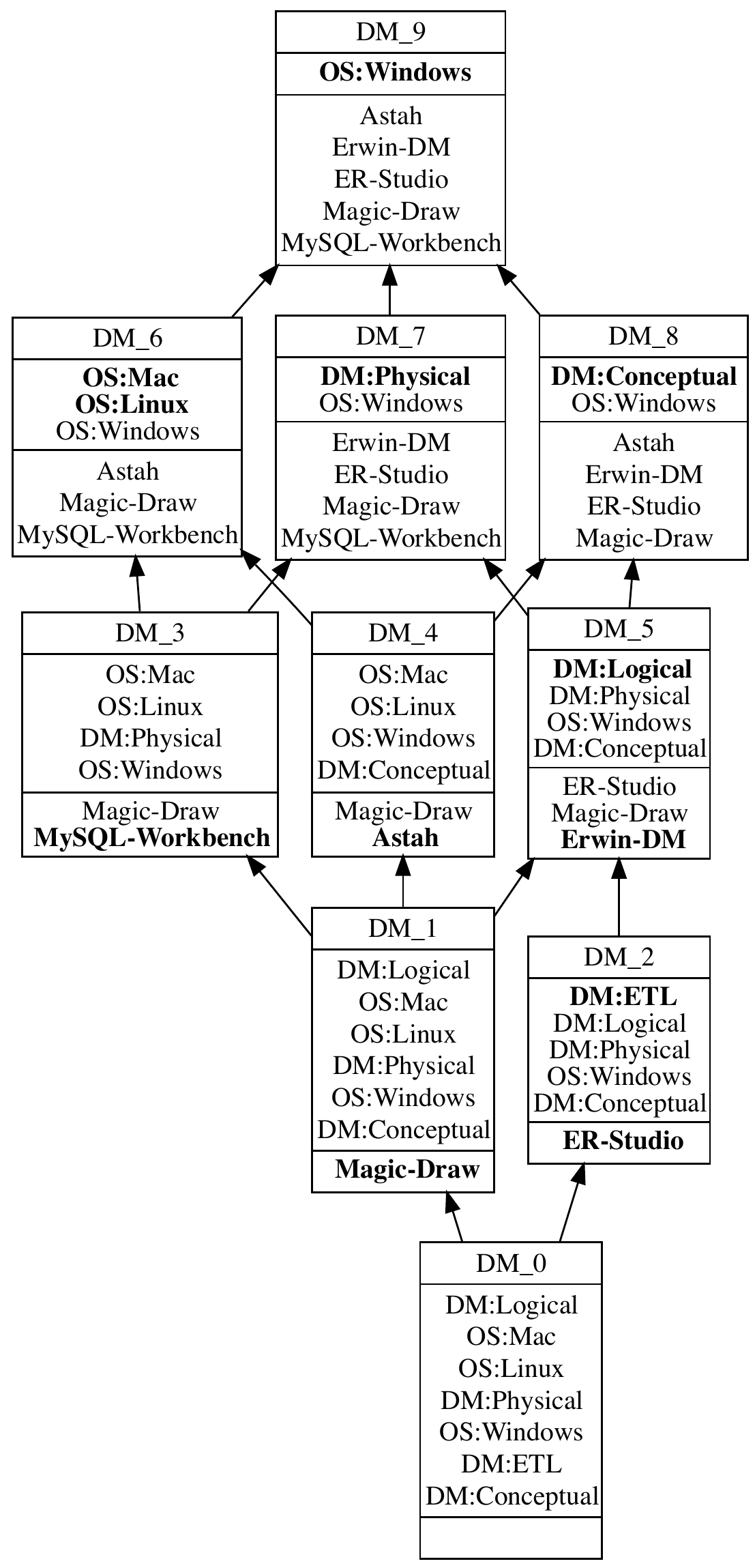}
    
    \caption{Concept lattice of $K_{DM}$ }
    \label{fig:lattice}
\end{figure}

The set of all identified formal concepts provided with the partial order $\leq$ forms a lattice structure called a concept lattice.
The concept lattice associated with the formal context of Table~\ref{tab:formalcontext} is presented in Fig.~\ref{fig:lattice}. 
A concept is represented by a 3-part box indicating the name of the concept (top part), its intent (middle part) and its extent (bottom part).
Arrows represent the partial order between the concepts: the arrow from $DM_3$ to $DM_6$ shows that $DM_3 \leq DM_6$. 
For the sake of readability, only the arrows corresponding to the transitive reduction are depicted.
The building process is deterministic and does not depend on any parameters, therefore:

\begin{property}
    The concept lattice is a canonical structure, i.e., only one concept lattice is associated with a given formal context.
\end{property}

A \textit{lattice} is a structure in which each subset of elements (in this case,  concepts) possesses at least one upper-bound and at least one lower-bound, inducing the "diamond shape" of the structure.
An \textbf{upper-bound} of a set of concepts $\{C_1, ..., C_n\}$ is a concept $C$ which is a super-concept of all of them, i.e., $C_i \leq C, \forall i \in [1..n]$.
Conversely, a \textbf{lower-bound} of a set of concept $\{C_1, ..., C_n\}$ is a concept which is a sub-concept of all of them, i.e., $C \leq C_i, \forall i \in [1..n]$.

%The set of all closed elements which can be extracted from $A$ form a closure system $(A, \alpha \circ \beta)$. Correspond to all intents of the concepts. It is stable by intersection and when associated with an inclusion relation, form a lattice structure.

\section{Variability information}
\label{sec:properties}

In the concept lattice, both the formal concepts and the specialization order highlight information which are true for the set of input objects.
In this section, we show how to read different types of information which are related to the variability of these objects.
We analyze which information we can infer from formal concepts, from the place of formal concepts in the structure and from how concepts are situated with respect to each other.
Then, we leverage this information to show that the lattice structure represents a space of decisions revealing variation points in existing configurations. 

\subsection{Formal Concepts}
We first study specific concepts called introducer concepts, and then discuss the parallel between concepts,  valid configurations and partial configurations.

\subsubsection{Introducer concepts}

By construction, a formal concept is a maximal set of objects sharing a maximal set of attributes. The three following statements are equivalent:

$\bullet$ there exists one concept in the concept lattice such that this concept has an attribute $a$ in its intent and that there is no larger subset of objects than its extent which all share $a$;

$\bullet$ this concept is the highest concept in the hierarchy which possesses $a$ in its intent;

$\bullet$ we say that this concept \textit{introduces} the attribute $a$: it is called an \textbf{attribute-concept}.

Indeed, there exists only one concept $C = (E, I)$ such that $E = \beta(\{a\})$ represents all objects having the attribute $a$. 
There cannot be a larger subset of objects than $E$ which  own $a$.
Moreover, because the concepts are organized depending on the set inclusion on their extents, the concepts having the largest extents are positioned in the highest part of the structure. 
The concept $C = ( \beta(\{a\}) , \alpha(\beta(\{a\})))$ is thus the highest concept of the structure having $a$ in its intent. 

\begin{property}
The extent of a concept introducing an attribute regroups exactly all objects possessing the introduced attribute.
\end{property}

In Fig.~\ref{fig:lattice},  $DM_7$ introduces the attribute \texttt{DM:Physical} and its extent corresponds exactly to the four objects having this attribute.

In a dual manner, the three following statements are equivalents:

$\bullet$ there exists one concept in the concept lattice such that this concept has an object $o$ in its extent and that there is no larger subset of attributes than its intent describing the object $o$;

$\bullet$ this concept is the lowest concept in the hierarchy which possesses $o$ in its extent;

$\bullet$ we say that this concept \textit{introduces} the object $o$: it is called an \textbf{object-concept}.

There is only one concept $C = (E, I)$ such that  $I = \alpha(\{o\})$ represents all attributes possessed by $o$.
There cannot be a larger set of attributes than $I$ which are owned by $o$.
Because the concepts are organized by reverse inclusion on their intents, the concepts with the largest intent are situated in the lowest part of the structure.
The concept $C = ( \beta( \alpha (\{o\})) , \alpha(\{o\}))$ is thus the lowest concept of the lattice having $o$ in its extent.

\begin{property}
The intent of a concept introducing an object regroups the exact set of attributes owned by the introduced object. It is thus representing a valid configuration of the formal context.
\end{property}

In Fig.~\ref{fig:lattice}, $DM_4$ introduces the object $Astah$ and its intent corresponds exactly to the four attributes owned by this object.

\begin{figure}[ht]
    \centering
    \includegraphics[width=.7\linewidth]{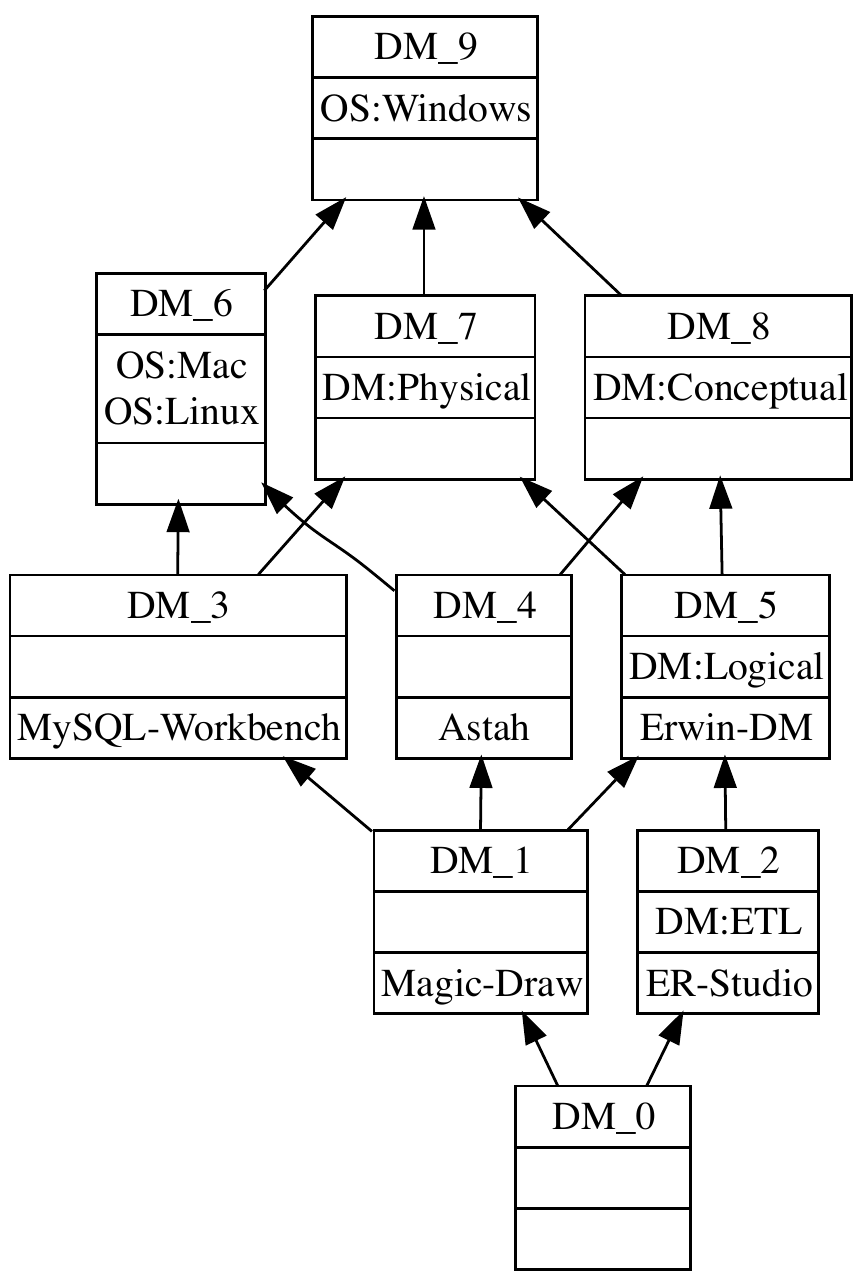}
    
    \caption{Concept lattice of $K_{DM}$ with simplified intents and extents}
    \label{fig:lattice-simplified}
\end{figure}

Thus, all concepts \textit{inherit} the attributes from their super-concepts and the objects from their sub-concepts.
This enable to present the concept lattice in a simplified way, by removing inherited attributes and objects, and only display them in the concepts they are introduced. 
The full intent and extent can then be reconstituted by inheritance. 
Figure~\ref{fig:lattice-simplified} presents the lattice of Fig.~\ref{fig:lattice} after simplification of its intents and extents.

\subsubsection{Partial Configurations}

We call an \textit{invalid configuration} a forbidden combination of attributes, such that there is no sequence of addition of attributes which leads to a valid configuration of the formal context. 
For instance, the subset $\{$\texttt{OS:Linux}, \texttt{DM:ETL}$\}$ is invalid because in the set of studied configurations, only \textit{ER-studio} owns \texttt{DM:ETL}, and this object does not have the attribute \texttt{OS:Linux}. 

We call a \textbf{partial configuration} a configuration which is not necessarily valid (i.e., it does not correspond to an object of the context), but which can become valid if we add the appropriate attributes.
For instance, $\{$\texttt{OS:Windows}, \texttt{OS:Mac}$\}$ is not a valid configuration of the context, but if we add \texttt{OS:Linux} and \texttt{DM:Conceptual}, we obtain the configuration of \textit{Astah}.
Note that adding \texttt{OS:Linux} and \texttt{DM:Physical} to the partial configuration leads to the configuration corresponding of \textit{MySQL-Workbench}.
A partial configuration can thus be shared by several valid configurations.

Concepts' intents represent subsets of attributes which are shared by objects: by construction, they do not represent forbidden combinations of attributes, and thus invalid configurations. 

\begin{property}
The intent of a concept introducing an object represents a valid configuration, and the intent of a concept which does not introduce an object represents a partial configuration.
\end{property}

Additionally, we saw that formal concepts represent maximal subsets of attributes:  partial configurations represented in concepts are thus \textit{maximal partial configurations}.
For instance, we saw previously that $\{$\texttt{OS:Windows}, \texttt{OS:Mac}$\}$ is a partial configuration.
However, it is not maximal: in the formal context, we can see that \texttt{OS:Mac} and \texttt{OS:Linux} always appear together in all objects. $\{$\texttt{OS:Windows}, \texttt{OS:Mac}, \texttt{OS:Linux}$\}$ is now a maximal partial configuration, which corresponds to concept $DM_6$.
Finally, because all maximal sets of objects and attributes are identified by mean of FCA, then:

\begin{property}
The set of concepts exactly represent all configurations and all maximal partial configurations from the formal context.
\end{property}

%Note that a partial configuration can be a valid configuration as well, e.g.,  when an object has all attributes of another object, plus other attributes.
%For instance, \textit{Magic-Draw} has all attributes of \textit{Astah}, plus \texttt{DM:Physical} and \texttt{DM-Logical}. 
%The configuration corresponding to \textit{Astah} is thus also a partial configuration found in \textit{Magic-Draw}.

\subsection{How Concepts are Situated in the Structure}

The specialization order naturally puts the concepts with the largest extents at the top of the conceptual structure, and the ones with the largest intents at the bottom.
The location of introducer concepts in the structure enables to estimate variability information regarding the attributes and/or objects they introduce.

Two concepts with a unique place in the concept lattice are particularly interesting.
The \textbf{top-concept} is the upper-bound of all concepts: it introduces attributes only if they are present in all objects, and they are thus inherited by all concepts. 
\begin{property}
The attributes which are present in all configurations (called \textit{core-features} in a SPL setting) are introduced in the top-concept.
\end{property}

The \textbf{bottom-concept} is the lower-bound of all concepts: if it introduces attributes but that its extent is empty, it means that these attributes are present in none of the objects. 
\begin{property}
If there exists attributes present in none of the configurations (called dead-features in a SPL setting), they are introduced in the bottom-concept whose extent is then empty.
\end{property}

If the bottom-concept introduces an object, then this object possesses all attributes of the formal context and there is no mutually exclusive attributes.

Attribute-concepts near the top of the structure usually possess more sub-concepts: attributes introduced in these concepts are thus shared by more (partial-)configurations and are more common or generic. 
Attributes introduced in concepts near the bottom of the structure are shared by less objects, and can thus be considered rarer or more specific.
Objects introduced in concepts closed to the top of the lattice have configurations that can be considered as more generic because they usually contain less attributes than objects introduced lower in the structure.
These more generic objects will possess more common attributes, because they are introduced higher in the structure.

\subsection{How Concepts Are Situated With Respect To Each Other}

A concept lattice is a partial order, thus all concepts are not comparable. 
%Comparable concepts are concepts which are linked by arrows in Fig.~\ref{fig:lattice}.
We call a \textit{chain} a sequence of concepts which are all comparable to each other. We can find a chain by following the arrows in concept lattices such as in Fig.~\ref{fig:lattice}.
For instance, the five concepts $DM_9$, $DM_6$, $DM_3$, $DM_1$ and $DM_0$ form a chain. If we add $DM_4$ to this set, it is not a chain anymore because $DM_3$ and $DM_4$ are not comparable.

We can assess the similarity of two comparable concepts depending on how they are situated with respect to each other in the hierarchy.
Let us consider three comparable and distinct concepts $C_1 = (E_1, I_1)$, $C_2 = (E_2, I_2)$ and $C_3 = (E_3, I_3)$.
If $C_1 < C_2 < C_3$, then $E_1 \subset E_2 \subset E_3$ and thus the difference between $E_1$ and $E_3$ is greater than the one between $E_1$ and $E_2$.
Therefore, the closer concepts are to each other in the structure, the more their extents are similar:

\begin{property}
    The most similar concepts of a given concept are its direct super-concepts and sub-concepts in the lattice; this set of concepts is called the \textbf{conceptual neighbourhood} of a concept.
\end{property}

This is again analogous to a class hierarchy, where the direct specializations of a given class are its most similar sub-classes.

The place of attribute-concepts with respect to each other provides information about constraints between attributes which are true in the considered formal context.
As we have seen before, an attribute-concept is a concept which introduces an attribute and its extent represents the group of all objects having this attribute.
Consider two comparable attribute-concepts $AC_1$ and $AC_2$ introducing respectively attributes $a_1$ and $a_2$, and that $AC_1$ is situated below $AC_2$ in a chain (i.e., $AC_1 \leq AC_2$).
We saw that $AC_1 \leq AC_2 \Leftrightarrow E_1 \subseteq E_2$, meaning that the set of all objects possessing $a_1$ is included in the set of all objects possessing $a_2$.
Therefore, for the considered input, all objects having $a_1$ also have $a_2$, and the binary implication $a_1 \rightarrow a_2$ then holds. 
In Fig.~\ref{fig:lattice}, because $DM_5 \leq DM_8$, we can extract the implication \texttt{DM:logical} $\rightarrow$ \texttt{DM:Conceptual}.
Thanks to Ganter and Wille~\cite{ganter2012formal} who showed that $AC_1 \leq AC_2 \Leftrightarrow a_1 \rightarrow a_2$, we can ensure that:

$\bullet$ All binary implications that are true in a formal context can be extracted from the associated concept lattice, making this extraction process \textbf{complete}.

$\bullet$ All binary implications that can be extracted from the concept lattice are true in its associated formal context, making this extraction process \textbf{sound}.

There is actually an equivalence between a formal context, its associated concept lattice, and the set of binary implications which can be identified in it. There is a large body of literature exploring the connection between binary implications and FCA, notably for building lattices of attribute sets from sets of implications as alternative to formal context~\cite{bertet2018lattices}.

Note that, if  $a_1$ and $a_2$ are introduced in the same concept (e.g., $DM_6$), then we have the double implication $a_1 \leftrightarrow a_2$.

The place of object-concepts with respect to each other provides information regarding how configurations are included in each other. 
Consider two object-concepts $OC_1$ and $OC_2$ introducing respectively the objects $o_1$ and $o_2$, and $OC_1 \leq OC_2$ ($OC_1$ is below $OC_2$ in a chain).
We have seen that in this case, $I_2 \subseteq I_1$ meaning that the configuration of $o_2$ is strictly included in the configuration of $o_1$.
In other words, $o_1$ has the same attributes as $o_2$, plus additional attributes: we can say that $o_1$ specializes $o_2$ or that $o_2$ generalizes $o_1$.
In Fig.~\ref{fig:lattice}, because $DM_2 \leq DM_5$, we can say that the configuration of \textit{ER-Studio} is a specialization of the configuration of \textit{Erwin-DM}.

We have seen in Sect.~\ref{sec:definitions} that a lattice is a structure in which all the sets of elements have at least one upper bound and one lower bound.
%An upper bound of a set of concepts is a concept which is greater than all concepts from the set. 
A set of concepts can have several upper bounds: in Fig~\ref{fig:lattice}, $DM_5$, $DM_7$, $DM_8$ and $DM_9$ are all upper bounds of $\{DM_1, DM_2\}$.
However, there is only one \textbf{lowest upper bound}: in our example, it is $DM_5$.
Conversely, a set of concepts can have several lower bounds, but only one \textbf{greatest lower bound}.
The lowest upper bound of two concepts provides interesting information regarding their similarity. 
Indeed, the \textbf{intent of the lowest upper bound corresponds exactly to the intersection of the intent of the two concepts}.
If we consider the lowest upper bound of two object-concepts (for instance, $DM_3$ and $DM_4$ introducing respectively \texttt{MySQL-Workbench} and \texttt{Astah}), the intent of this concept represents the set of all attributes which are present in both configurations, i.e., representing their similarity.
For $DM_3$ and $DM_4$, the lowest upper bound is $DM_6$, showing that they have in common the fact that they support the three OS \texttt{Windows}, \texttt{Mac} and \texttt{Linux}, but no data models.
The intent of the \textbf{greatest lower bound of two concepts} includes  the attributes of their intents, however, it does not always represent the exact union of their intents: it \textbf{represents the smallest maximal group of objects described by the attributes of both concepts' intent}. 
If the greatest lower bound of two object-concepts is also an object-concept, it means that the latter possesses the attributes of both objects.
In Fig.~\ref{fig:lattice}, the greatest lower bound of $DM_4$ (introducing \textit{Astah}) and $DM_5$ (introducing \textit{Erwin-DM}) is $DM_1$ (introducing \textit{Magic-Draw}).
Magic draw thus possesses all features of both \textit{Astah} and \textit{Erwin-DM}.
If the greatest lower bound of two attribute-concepts is a concept with an empty extent (i.e., the bottom-concept), it means that there is no object having both these attributes in the considered formal context.
In other words, the two attributes are \textbf{mutually exclusive}.
For instance, the greatest lower bound of $DM_6$ and $DM_2$ is $DM_0$, whose extent is empty: thus, \texttt{OS:Linux} $\rightarrow \lnot$ \texttt{DM:ETL}.

\subsection{Fitting all Properties Together: a Concise Browsable Space of Decisions}

We have seen that the largest clusters are situated at the top of the structure, the top-concept representing the cluster of all objects. 
Moreover, the conceptual neighbourhood of a concept regroups the concepts which are the most similar to it.
Thus, the concepts directly under the top-concept are the largest maximal clusters representing a subdivision of its extent according to objects' similarity.
Each level of the structure thus subdivides the concepts of the upper-level in concepts of thinner granularity while ensuring their maximality. 
\begin{property}
    The concept lattice organizes clusters by refinement at several levels of increasing details, starting from the clusters of all objects at the top towards the clusters of one object at the bottom.
\end{property}

Thanks to this property, concept lattices have been considered great candidates to support \textit{exploratory search}~\cite{godin1986lattice}, an information retrieval strategy tailored to situations where the user may not be able to query documents based on their associated keywords~\cite{marchionini2006exploratory,white2009exploratory}.
Instead, it suggests to browse through the collection of documents step-by-step, allowing users to gradually discover and comprehend the existing documents and their keywords.

Concept lattices offer a structured space of navigation, enabling to progressively explore groups of similar objects and their shared attributes by navigating from concepts to concepts, a task refered to as \textbf{conceptual navigation}~\cite{codocedo2015formal}.
For that, the user is always placed on one concept (the \textit{current concept}) and can "move" to a close concept, which thus becomes the new current concept.
Because the intent of a given concept  strictly includes the intents of its super-concepts, we can thus obtain the super-concepts of the current concept by \textit{deleting attributes} from its intent.
In the lattice, super-concepts thus show all maximal partial configurations which are more generic than the one of the current concept.
They represent the smallest sequence of attribute deletion to obtain maximal partial configurations.
In other word, going from the current concept to one of its super-concepts thus represent the minimal possible decisions to reduce the current set of attributes without breaking the property of maximality.

For instance, let us consider the concept $DM_4$ as the current concept, whose extent includes the two objects \textit{Astah} and \textit{Magic-Draw}.
It has two direct super-concepts, namely $DM_6$ and $DM_8$.
Note that $DM_9$ is also one of its super-concepts, but not a direct one. 
From the point of view of $DM_4$, moving to $DM_6$ represents the deletion of \texttt{DM:Conceptual}.
This decision enables to consider \textit{MySQL-Workbench} in the set of objects matching the obtained partial configuration.
Moving to $DM_8$ represents the deletion of \texttt{OS:Mac} and \texttt{OS:Linux}, which enables to include \textit{Erwin-DM} and \textit{ER-Studio} in the set of objects sharing this partial configuration.
A super-concept which is not direct impact more strongly the set of selected attributes.
Going from $DM_4$ to $DM_9$ would remove \texttt{DM:Conceptual}, \texttt{OS:Mac} and \texttt{OS:Linux} in one go, which is not minimal.

In a dual manner, the intent of the current concept is strictly included in the ones of its direct sub-concepts: sub-concepts are thus obtained by \textit{adding attributes} to the current intent.
They thus represent the closest more specific partial configurations of the current concept.
The direct sub-concepts represent the smallest sequence of addition of attributes to obtain maximal partial configurations.

\begin{property}
    Following a chain top-down represents a sequence of attribute addition by minimal steps, while following a chain bottom-up shows a sequence of attribute deletion by minimal steps.
\end{property}

Transitions between concepts can thus be seen as \textbf{decision nodes in the process of selecting attributes to obtain a valid configuration}, where different sequences of addition diverge towards different valid configurations or larger partial configurations.
Godin et al.~\cite{godin1986lattice} defined these decisions as minimal, meaning that there are the smallest set of decisions which can be made in this space of configurations. 
Concept lattices thus represent a space of decisions by exposing variation points in existing configurations. 
\section{Exploiting conceptual structures}
\label{sec:extensions}

In this section, we discuss how practitioners may customize the FCA framework to suit their needs. 
We introduce methods to condense concept lattices and address scalability concerns. 
Then, we present existing FCA extensions that facilitate the handling of complex inputs.
Finally, we provide few pointers on available tools.

\subsection{Sub-Hierarchies}

\begin{figure}
    \centering
        \includegraphics[width=.7\linewidth]{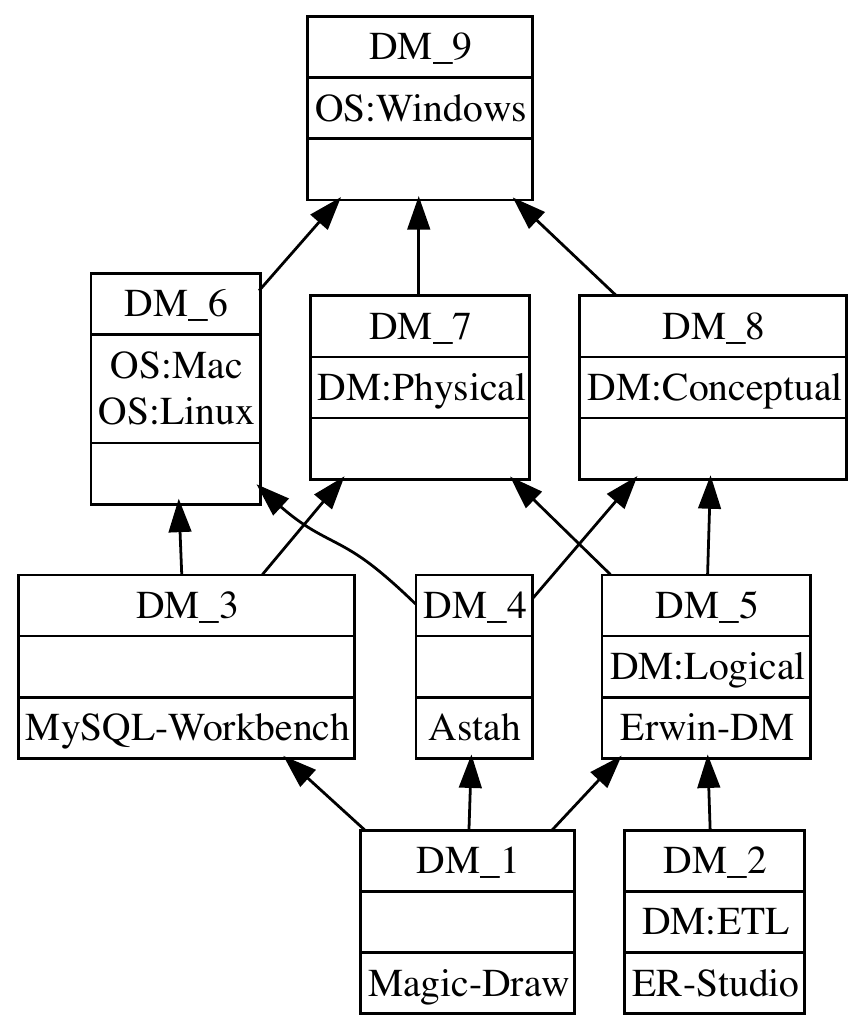}
    \caption{AOC-poset of $K_{DM}$}
    \label{fig:aoc-poset}
\end{figure}

\begin{figure}
    \centering
        \includegraphics[width=.8\linewidth]{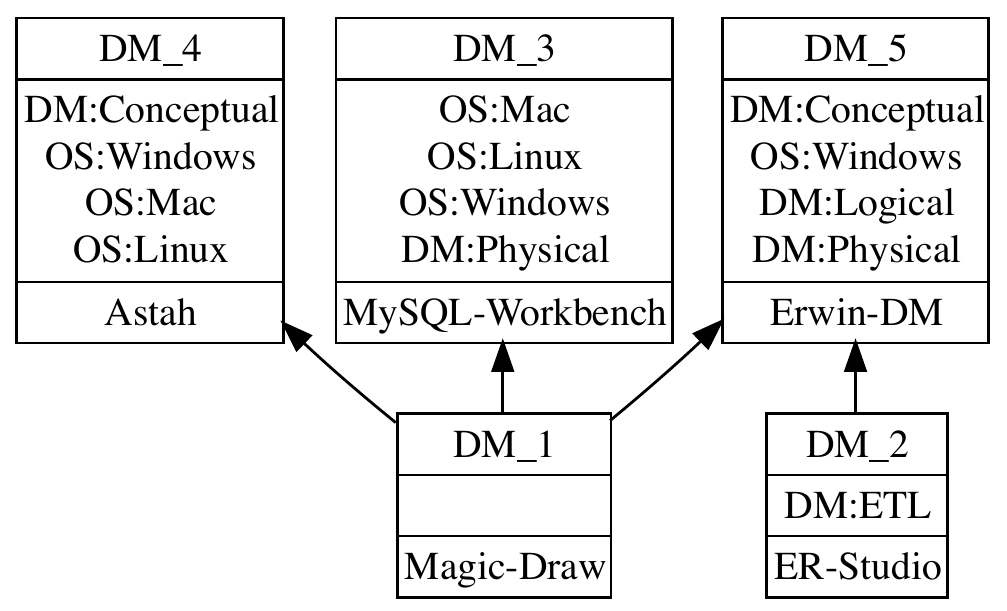}
    \caption{OC-poset of $K_{DM}$}
    \label{fig:oc-poset}
\end{figure}

\begin{figure}
    \centering

        \includegraphics[width=.8\linewidth]{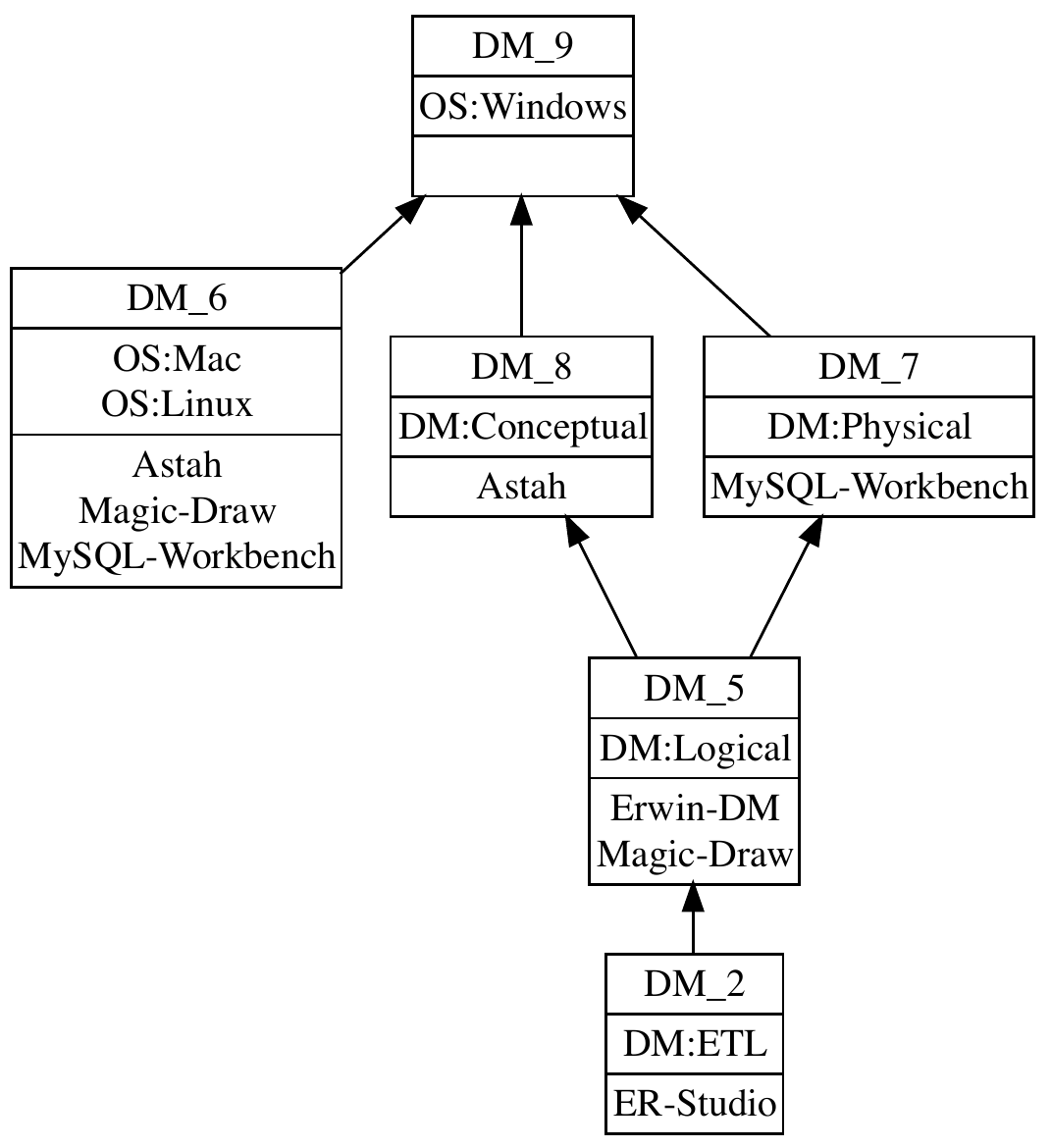}

    \caption{AC-poset of $K_{DM}$}
    \label{fig:ac-poset}
\end{figure}

In some applications, only a subset of the concepts are of interest and thus building the full lattice is not necessary.
For instance, when studying the organization of introducer concepts to extract binary implications.
In that case, it is possible to build a sub-hierarchy restricted to some of its concepts. There are three well-known sub-hierarchies: the AOC-poset (for \textit{Attribute- and Object-Concepts partially ordered set}) gathers all introducer concepts, the AC-poset (\textit{Attribute-Concepts poset}) retains only the concepts introducing attributes and the OC-poset (\textit{Object-Concepts poset}) only the ones introducing objects. 
Figure~\ref{fig:aoc-poset} presents the AOC-poset of Table~\ref{tab:formalcontext}, with simplified intent.
Figures~\ref{fig:oc-poset} and~\ref{fig:ac-poset} represent the OC-poset and AC-poset, respectively.
Note that these sub-hierarchies are not lattice structures, and some of the previous properties which are true in a concept lattice do not hold in sub-hierarchies.
These three sub-hierarchies are canonical structures (e.g., only one AOC-poset can be built from a given context). 
All groups of concepts do not necessarily have an upper or lower bound.
In the AOC and AC-poset, if there exist attributes present in all objects, the sub-hierarchies will have a top-concept introducing them. Same applies for attributes present in none of the objects and the bottom-concept.
Conceptual neighbourhood in sub-hierarchies also ensures to show the closest concepts.

For applications which seek to identify a limited set of relevant clusters, the concepts situated close to the top of the lattice are more relevant.
They represent large groups of objects described by common attributes and their granularity gets thinner the lower we go down in the structure.
In these situations, generating only the top of the structure (called the \textbf{iceberg-lattice}~\cite{stumme2002computing}) is enough. 
If one wants to retrieve only the clusters grouping at least $n$ objects (i.e., the concepts with an extent's size of $n$ or more), they correspond to a portion of the top of the lattice: we can thus prune all chains starting from the concept whose extent's size is less than $n$.
The larger the $n$, the smaller the portion retrieved.

\subsection{FCA's Extensions}

The theory and application of FCA do not depend on the input type, but rather on properties it verifies.
This contributes to the framework's flexibility, leading to  numerous extensions considering diverse input types (see Table~\ref{tab:extensions}).

%\paragraph{Pattern Structures}

Among the proposed extensions, the \textbf{Pattern Structures}~\cite{ganter2001pattern} (PS) framework has a particular importance.
It presents a \textit{generalization of FCA} which has the potential to consider objects associated with any type of descriptions, and not only sets of binary attributes.

PS defines the conditions that an input (of the form \textit{objects} $\mapsto$ \textit{descriptions}) has to verify to enable the creation of a concept lattice: i) all descriptions should be of the same type, ii) one should be able to characterize the similarity of a sub-set of these descriptions, and iii) the characterized similarity should be of the same type as the  descriptions.
In other words, a set of descriptions $D$ of the same type \texttt{t} should be associated with a similarity operator $\sqcap$, such that when provided with $n$ descriptions of type \texttt{t}, it returns a description of type \texttt{t} representing their similarity.
For instance, if objects are described by interval of values, we can define a similarity operator returning the interval corresponding to their intersection:  $\sqcap_{inter}$ such that $[a_1, b_1]\sqcap_{inter} [a_2, b_2] = [max(a_1, a_2), min(b_1, b_2)]$.
The type of descriptions and the similarity operator are only constrained by these properties, and we can thus define different similarity operators over the same type of descriptions.

($D, \sqcap$) forms a meet-semilattice, i.e., a structure in which each subset of descriptions from $D$ has an upper bound, which is the element in $D$ that depicts their similarity.

Pattern Structures then proposes a re-definition of the two operators $\alpha$ and $\beta$ so they can be applied on any set of descriptions meeting these conditions. 
%With $O$ the set of objects and $D$ the set of descriptions, we have $\alpha : 2^O \mapsto D$ and $\beta: D \mapsto 2^O$
%The operator $\alpha$ associates with a subset of objects $O' \in O$ the most specific description of $(D, \sqcap)$ matching all objects of $O'$. The operator $\beta$ associates a description $d \in (D, \sqcap)$ with all objects from $O$ matching it.
The application of the two operators enables to extract a finite set of concepts (called \textit{pattern concepts}): a pattern concept is a bi-cluster representing a maximal set of objects (extent) with the most specific description characterizing all of their similarity (intent).
They can be organized by specialization to obtain a lattice structure (\textit{pattern concept lattice}) having the same properties regarding variability which are presented in Section~\ref{sec:properties}.

If we consider that the descriptions are sets of binary attributes, and that the similarity between two sets of binary attributes is characterized by their intersection, then we fall back in the traditional case of FCA.

Some other extensions do not maintain all properties of FCA but only a certain number of them, which enables more flexibility than PS on the form of the handled input. 
That's why other extensions exist despite that PS generalizes FCA to any kind of descriptions.
For instance, Triadic Concept Analysis is an extension considering ternary relationships between objects, their descriptions and a third set usually referred as "conditions". 
It can process input of the form \textit{object o is associated with description d under condition c}, which is not possible with the Pattern Structures framework.
Also, even when some types of inputs can be covered with PS, some dedicated extensions can help ease their processing and comprehension by providing dedicated re-definitions and tools.

\begin{table}[ht]
\caption{Prevalent extensions of Formal Concept Analysis}
\label{tab:extensions}

\small
\begin{tabular}{|l|l|c|}
\hline
\textbf{Input} & \textbf{Extension's name and references} \\ \hline
Descriptions $(D, \sqcap)$              & Pattern Structures: \cite{ganter2001pattern}           \\ \hline
Relational data         & Relational Concept Analysis: \cite{rouane2013relational}\\ \hline
Temporal data           & Temporal Concept Analysis: \cite{neouchi2001towards,wolff2002interpretation} \\ \hline
Logic formulas           & Logical Concept Analysis: \cite{ferre2000logical}           \\ \hline
Knowledge graphs                  &Graph-FCA: \cite{ferre2020graph}            \\ \hline
Multi-dimensional data  & Polyadic / Triadic Concept Analysis: \cite{voutsadakis2002polyadic,ignatov2015triadic}          \\ \hline
Fuzzy   contexts                & Fuzzy Concept Analysis:~\cite{buelohlavek2004concept}           \\ \hline
\end{tabular}
\end{table}

Table~\ref{tab:extensions} presents some of the most prevalent extensions of FCA found in the literature, as well as the input they can process.
\textbf{Relation Concept Analysis} enables to consider several formal contexts (such as the one in Table~\ref{tab:formalcontext}) as well as relationships between objects of different contexts (e.g., between data modelling tools and a potential other context describing software licences and their characteristics).
In the resulting concept lattices, objects are both organized depending on the attributes they share, but also depending on objects (from other concept lattices) they are related to.
\textbf{Temporal Concept Analysis} applies FCA to temporal data. It considers formal contexts of objects whose descriptions are associated with temporal aspects. It enables to represent and extract knowledge including a temporal dimension, and thus to analyze datasets which are varying over time.
\textbf{Logical Concept Analysis} considers objects which are described by formulas of a formal logic. It maintains all properties of FCA, provided that these formulas are associated with a disjunctive ($\lor$) and conjunctive ($\land$) operations.
\textbf{Graph-FCA} is an extension for handling knowledge graphs such as conceptual graphs or RDF graphs. It enables considering complex data in the form of objects interlinked by binary and n-ary relationships.
\textbf{Polyadic Concept Analysis} enable to consider n-ary relationships between elements of $n$ dimensions. The tridimensional case (\textbf{Triadic Concept Analysis}, linking objects, descriptions and conditions) mentioned at the beginning of this section is the most prominent use-case.
Finally, \textbf{Fuzzy Concept Analysis} considers  uncertainty regarding if an object owns a certain attribute or not. It relies on a fuzzy formal context associating each $(o, a) \in R$ with a value between 0 and 1 estimating the degree of truth for the assertion "the object \textit{o} possesses the attribute \textit{a}".

Note that the objective of this section is not to offer a comprehensive introduction to each FCA's extension. 
It rather shows that FCA demonstrates a powerful flexibility both on the input which can be considered and the knowledge which can be extracted. 
Even if all FCA properties which are useful for variability representation are not maintained in some of these extensions, the generated concepts and their hierarchy can always support the analysis of at least some of the previously defined variability information.

\subsection{Available Tools}
Since 2007, Uta Priss is maintaining a list of tools and software for FCA\footnote{\url{https://upriss.github.io/fca/fcasoftware.html}} well-known in the community.
These tools are predominantly academic tools tailored for a specific task or extension (e.g., fcaR\footnote{\url{https://github.com/neuroimaginador/fcaR}}, a R package for fuzzy concept analysis, RCAExplore~\cite{dolques2019rcaexplore}, a standalone software to handle relational datasets).

The Python library \texttt{concepts.py}\footnote{\url{https://pypi.org/project/concepts/}} proposes methods to create contexts, build formal concepts and concept lattice as well as visualize the final structure with \texttt{graphviz}.

\texttt{GALACTIC} (for \textit{Galois Lattices,Concept Theory, Implicational systems and Closures})~\cite{demko2022galactic} is a collection of Python 3 packages generalizing FCA algorithms to heterogeneous and complex data, built upon  Pattern Structures and Logical Concept Analysis.
\section{Related Work}
\label{sec:rw}

Formal concept analysis and conceptual structures were used several time for variability modeling and management. 

Loesch and Ploedereder~\cite{DBLP:conf/csmr/LoeschP07} built a concept lattice from the list of valid configurations of a feature model.
They then relied on the obtained structure to extract core and dead features, mutex, co-occurences, and use the tool ConExp to derive implications.
Because the concept lattice naturally represents the variability of the set of configurations, they used the extracted information to restructure the FM such that it reflects more closely the variability of its configurations.
Similarly, Carbonnel et al.~\cite{DBLP:conf/enase/CarbonnelHMN17} built a formal contexts merging configuration sets from different variability models to synthesize a feature model representing their union or intersection.

Rather than analyzing variability of existing variability models, a large body of work aimed at creating FMs based on variability information extracted from descriptions of software variants.
The used descriptions are diverse: data access semantics~\cite{DBLP:conf/wcre/YangPZ99a}, function blocks (encapsulated algorithms)~\cite{DBLP:conf/csmr/LoeschP07},  propositional formulas~\cite{DBLP:journals/infsof/SheRAWC14}, set of features~\cite{DBLP:conf/cla/Al-MsiedeenHSUV14,DBLP:journals/jss/CarbonnelHN19}, functional requirements~\cite{DBLP:journals/cj/MeftehBB16}, user-stories~\cite{DBLP:conf/vamos/GeorgesRHKNT23} or even agile specifications~\cite{DBLP:conf/splc/Georges22}.
This demonstrates the adaptability of the FCA framework to various kinds of inputs and its capacity to identify variability information in a wide range of scenarios.

Shatnawi et al.~\cite{DBLP:journals/jss/ShatnawiSS17} used FCA to identify commonalities and variabilities within sets of software components to derive, not a feature model, but a software product line architecture.
Niu and Easterbrook~\cite{DBLP:conf/aosd/NiuE09} used concept lattices to obtain insights on the variability of requirements in different quality scenarios, which enabled to find functional units, detect potential interactions between requirements and  analyze the impact of modifying requirements.

We can also find several works using FCA for feature location.
Eisenbarth et al.~\cite{DBLP:journals/tse/EisenbarthKS03} used FCA to map features to their associated code in a single software product.
Their approach relies on the definition of scenarios invoking features, and the identification of the computation units which are activated when running each scenario.
Concept lattices are built from this input to cluster computational units depending on the scenarios which activate them.
They analyzed the obtained structure to derive information regarding computational units variability, e.g., units ran by all scenarios, units specific to one scenario, scenario activating all units. 
They also analyzed the place of the units in the lattice to assess their level of specificity.
Provided with the information of which features are invoked by which scenario, they extract a mapping between features and computational units.
Here again the flexibility of formal contexts and the numerous information emphasized by conceptual structures enabled variability analysis on diverse software artifacts.
Xue et al.~\cite{DBLP:conf/wcre/XueXJ12} relied on commonalities detected by mean of FCA in features and implementations of several variants to improve the performance of information retrieval techniques used for feature location.
Hlad et al. completed the previous approach by leveraging relational concept analysis to consider feature interactions~\cite{DBLP:conf/gpce/HladLHS21}.

Other works relied on FCA's extensions to consider inputs which could not be easily captured in formal contexts.
In~\cite{DBLP:journals/jss/CarbonnelHN19a}, the authors used Pattern Structures to extract extended feature models (including attributes, cardinalities) from non-bolean formal contexts.
Pattern Structures were also used  to characterize the variability of similar code snippets based on their abstract syntax tree~\cite{galasso2022fine}.
Bazin et al.~\cite{bazin2023} explored the ability of Triadic Concept Analysis to capture both variability in space and in time.

\section{Perspectives and Conclusion}
\label{sec:conclusion}

In this paper, we discussed the suitability of the Formal Concept Analysis framework for representing and analyzing variability of similar elements described by binary attributes.
We gathered definitions of the core concepts which can be of interest for software engineers, and presented a guide to read the diverse variability information naturally present in the built structures. 
Finally, we briefly presented sub-structures, extensions and tools which could be reused by the community in applications or to consider diverse types of inputs, hence helping to broaden the current scope variability extraction and management.

To further support practitioners  applying Formal Concept Analysis, we plan to extend this current work to study available extensions in more details through definitions, examples of artifacts which can be handled and examples of information which can be extracted. 
Another useful work would be to develop a dedicated library implementing variability extraction operations based on the properties discussed in Section~\ref{sec:properties}, which could leverage existing Python libraries for building and handling concept lattices.

\bibliographystyle{abbrv}
\bibliography{software.bib}
\end{document}